%% file: main.tex
\documentclass[a4paper]{article}
\usepackage{INTERSPEECH2022}

\usepackage{times}
\usepackage{booktabs}
\usepackage{amsmath}
\usepackage{amsfonts}
\usepackage{graphicx}
\usepackage{latexsym}
\usepackage{algorithm}
\usepackage{algpseudocode}
\usepackage{kotex}  
\usepackage{multirow}
\usepackage{caption}
\usepackage{subcaption}

\usepackage[T1]{fontenc}
\usepackage[utf8]{inputenc}

\usepackage{pifont}
\newcommand{\cmark}{\ding{51}}
\newcommand{\xmark}{\ding{55}}

\title{Korean Tokenization for Beam Search Rescoring in Speech Recognition}
\name{Kyuhong Shim, Hyewon Bae, Wonyong Sung}
\address{Department of Electrical and Computer Engineering, Seoul National University, Korea}
\email{\{skhu20, hyewon0309, wysung\}@snu.ac.kr}

\begin{document}
\maketitle

\begin{abstract}    
\input{section/abstract}
\end{abstract}

\input{section/introduction}
\input{section/background}
\input{section/method}
\input{section/experiment}
\input{section/conclusion}
\newpage
\bibliographystyle{IEEEtran}
\bibliography{reference}

\end{document}

%% file: section/abstract.tex
The performance of automatic speech recognition (ASR) models can be greatly improved by proper beam-search decoding with external language model (LM).
There has been an increasing interest in Korean speech recognition, but not many studies have been focused on the decoding procedure.
In this paper, we propose a Korean tokenization method for neural network-based LM used for Korean ASR.
Although the common approach is to use the same tokenization method for external LM as the ASR model, we show that it may not be the best choice for Korean.
We propose a new tokenization method that inserts a special token, SkipTC, when there is no trailing consonant in a Korean syllable.
By utilizing the proposed SkipTC token, the input sequence for LM becomes very regularly patterned so that the LM can better learn the linguistic characteristics.
Our experiments show that the proposed approach achieves a lower word error rate compared to the same LM model without SkipTC.
In addition, we are the first to report the ASR performance for the recently introduced large-scale 7,600h Korean speech dataset.

%% file: section/introduction.tex
\section{Introduction}\label{sec:intro}

For end-to-end automatic speech recognition (ASR) systems, language models (LMs) are widely used for reducing the error rate.
LMs are especially useful when the speech audio itself is ambiguous as to which sentence is pronounced.
External LM produces an additional likelihood score for the current beam during beam search decoding, where the score from the acoustic model (AM) and the LM are jointly considered.
To generate such scores, N-gram LMs~\cite{ngram} and deep neural network-based models are the most popularly applied ones.
Especially, the latter shows much better improvement in the error rate because of their powerful ability to model the likelihood of a given sentence.
For example, recurrent neural networks (RNNs), such as LSTM and GRU, and recently Transformer-based autoregressive LMs~\cite{TXL} have been widely adopted.
Extensive studies have been conducted to improve the quality of LM, mainly focused on exploiting more powerful architectures and training such models with a vast amount of data~\cite{rethinkingasr}.

Tokenization is the key ingredient for the performance of LM, yet their effect on the ASR decoding has not been much studied.
Tokenization splits a sentence into a sequence of linguistic tokens, where the tokens compose a vocabulary used for speech recognition or language modeling.
There exist word-based, character-based, and subword-based (e.g., byte-pair encoding \cite{BPE}) tokenization methods.
It is a common approach for recent ASR systems to employ character or subword as the minimum token unit to evade the out-of-vocabulary problem~\cite{BPE,unigram} and restrict the vocabulary size.
For the efficiency of the decoding, the same tokenization method is usually used for acoustic and language models, although exploiting the same vocabulary for ASR and LM is not always necessary.
By sharing the same tokenization method, the token boundary can be aligned during the decoding process and LM can provide the likelihood score for each left-to-right step.
On the other hand, two-pass rescoring~\cite{twopassfast,twopassprune} evaluates the complete sentences after initial beam search for the second pass so that the token boundary alignment is not very important.

Recently, studies on Korean speech recognition have been actively conducted~\cite{koreanclova,koreanopen}.
These advances in Korean ASR are mainly boosted by the release of large-scale Korean speech corpus, such as KSponSpeech~\cite{ksponspeech} which includes about 970h of conversational dialogues.
Several studies have achieved fairly good recognition accuracy in Korean ASR by adopting neural networks as AM~\cite{koreanctcatt,koreanw2v}.
However, the decoding process and corresponding LMs have not been very much studied for Korean ASR.
For example, external LM was not employed~\cite{ksponspeech,koreanctcatt} or only an N-gram LM was utilized~\cite{koreanextend}.
Because Korean utterances can be written in multiple forms when dictated as-is, the importance of LM rescoring is significant.

\begin{figure}[t]
    \vspace{0.3cm}
    \centering
    \resizebox{0.98\columnwidth}{!}{
    \includegraphics{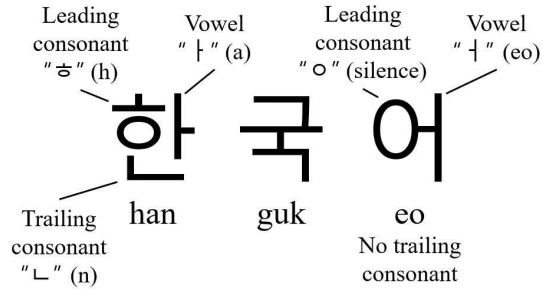}}
    \caption{Korean syllables and their components. Note that the trailing consonant can be skipped in the written form, as shown in the third syllable. The word `한국어' means `Korean.' }
    \label{fig:korean}
    \vspace{-0.2cm}
\end{figure}

In this paper, we propose a new Korean tokenization method that utilizes the characteristics of the Korean syllable composition for two-pass decoding of ASR with LM.
The key idea of the proposed method is to add an auxiliary token, to make the tokenized sequence more regular by forcing two types of tokens to appear alternately within a word.
Figure~\ref{fig:korean} shows the structure of Korean syllables (see Section~\ref{ssec:korean} for details).
We can observe that each Korean syllable can be separated into two components: (1) leading consonant + vowel and (2) trailing consonant, where the former always exists but the latter is optional.
During the tokenization, we insert the auxiliary token, named \textit{SkipTC} token, when there is no trailing consonant in a Korean syllable.
As a result, LM can always encounter alternating token types (1) and (2) and better learn the characteristics of Korean sentences.
We employ the proposed tokenization method for RNN-based LM, and then use the LM for the rescoring.

\begin{table*}[ht]
    \centering
    \caption{Comparison of different Korean tokenization methods. `` * '' symbol is the proposed \textit{SkipTC} token, which is used to indicate the trailing consonant is missing in a syllable. For each example, the first row demonstrates the tokenized Korean sentence, and the second row shows the pronunciation of each token.}
    \resizebox{0.92\linewidth}{!}{
    \begin{tabular}{ll|l}
        \toprule
        Tokenization & Token composition & Example \\
        \midrule
        \midrule
        \multirow {2} {*} {Syllable-based} & \multirow {2} {*} {LC + V + TC} 
                    & 나/ 는/  / 집/ 에/  / 간/ 다/ \\ 
                &   & na/ neun/  / jip/ e/  / gan/ da/ \\
        \midrule
        \midrule
        \multirow {2} {*} {Phoneme-based} & \multirow {2} {*} {LC, V, TC} 
                    & ㄴ/ ㅏ/ ㄴ/ ㅡ/ ㄴ/  / ㅈ/ ㅣ/ ㅂ/ ㅇ/ ㅔ/  / ㄱ/ ㅏ/ ㄴ/ ㄷ/ ㅏ/ \\ 
                &   & n/ a/ n/ eu/ n/  / j/ i/ p/ e/  / g/ a/ n/ d/ a/ \\
        \midrule
        \multirow {2} {*} {+ \textit{SkipTC}} & \multirow {2} {*} {LC, V, \{TC, SkipTC\}}
                    & ㄴ/ ㅏ/ */ ㄴ/ ㅡ/ ㄴ/  / ㅈ/ ㅣ/ ㅂ/ ㅇ/ ㅔ / */  / ㄱ/ ㅏ/ ㄴ/ ㄷ/ ㅏ/ */ \\ 
                &   & n/ a/ -/ n/ eu/ n/  / j/ i/ p/ e/ -/  / g/ a/ n/ d/ a/ -/ \\
        \midrule
        \midrule
        \multirow {2} {*} {Phoneme-based (ours)} & \multirow {2} {*} {LC + V, TC} 
                    & 나/ 느/ ㄴ/  / 지/ ㅂ/ 에/  / 가/ ㄴ/ 다/ \\ 
                &   & na/ neu/ n/  / ji/ p/ e/  / ga/ n/ da/ \\
        \midrule
        \multirow {2} {*} {+ \textit{SkipTC}} & \multirow {2} {*} {LC + V, \{TC, SkipTC\}} 
                    & 나/ */ 느/ ㄴ/  / 지/ ㅂ/ 에/ */  / 가/ ㄴ/ 다/ */ \\ 
                &   & na/ -/ neu/ n/  / ji/ p/ e/ -/  / ga/ n/ da/ -/ \\
        \bottomrule
    \end{tabular}}
    \label{tab:compare}
\end{table*}

Our experiments show that using the proposed SkipTC token for LM tokenization reduces the word error rate and the syllable error rate of Korean ASR by 2.68\% and 0.73\%, respectively.
We identify that the improvement comes from the enhanced ability of LM to model the likelihood, presented as the reduced perplexity.
We evaluate the proposed Korean tokenization method on the recently released large-scale 7,600h Korean speech dataset.
Please note that we are the first to train an ASR model and report the performance on this dataset.

%% file: section/background.tex
\section{Background and Related Work}\label{sec:related}

\subsection{Beam Search Decoding and RNN-LM}\label{ssec:rescoring}

The goal of speech recognition is to find the most probable transcription for a given utterance.
The ASR system transforms the input sequence of acoustic frames $X =\{x_1, x_2, ..., x_T\}$ of length $T$ to the output sequence of linguistic tokens $Y = \{y_1, y_2, ..., y_L\}$ of length $L$.
Formally, the decoding process searches the most probable transcription $Y^{*}$ that maximizes the score $S(Y)$:
\begin{align}
\vspace{-5pt}
Y^{*} &= \text{argmax}_Y S(Y) \\
S(Y)  &= \log(p_\text{AM}(Y|X)) + \alpha \log(p_\text{LM}(Y)) + \beta |Y|, \label{eq:score}
\end{align}
where $|Y|$ indicates the number of tokens within the sequence and $\alpha, \beta$ are weighting coefficients.
$\alpha$ is the LM weight that controls the contribution of LM, and $\beta$ is the insertion bonus that encourages more words in the transcription, following~\cite{deepspeech2,hwang}.
In practice, the above scoring is combined into the beam search decoding and used to find the topmost probable beam candidates for each left-to-right step.
Step-by-step beam search is possible because the LM is usually designed to produce the likelihood score of the prefix of a sentence, $p_{\{\text{LM}\}}(y_{1:i})$.

RNN-based language model (RNN-LM) is an autoregressive generative model that shows impressive performance on the probability density estimation of the given sequence~\cite{nnlm}.
RNN-LM factorizes the probability of the entire sequence by the chain rule as $p(Y)=\prod_{i=1}^{L} p(y_i|y_{1:i})$.
Each term $p(y_i|y_{1:i})$ is implemented as a single forward pass of RNN cell, such as LSTM or GRU, that takes a current token as input and outputs the probability distribution of the next token.
The power of RNN-LMs comes from their ability to learn long-term dependencies and patterns.
In particular, it is known that LSTM performs particularly well when dealing with problems related to regularly patterned input~\cite{lstmcount,lstmpattern}.

We briefly describe the beam search decoding process, following~\cite{hwang}.
The beam is a candidate of the final transcription where each beam consists of prefix string and score.
For each time step, each beam is extended by every possible token and generates new beam candidates.
Then, the scores $S(y_{1:i})$ of new candidates are calculated and sorted.
Usually, the topmost probable $B$ candidates are preserved while others are discarded, where $B$ is referred to as the beam width.
This pruning procedure is essential because maintaining every beam causes the exponential growth of computation and memory cost.
Extensive studies have been conducted~\cite{decodingcompare,decodingword} to improve the performance and efficiency of beam search decoding.
For example, subword-based decoding~\cite{decodingsubword}, blockwise decoding~\cite{decodingblockwise}, and bidirectional LM decoding~\cite{decodingfuture} have been proposed.
On the other hand, two-pass rescoring is also widely used~\cite{twopasse2e,twopassprune}; usually, the first pass calculates the beam candidates with fast LM, such as N-gram LM, and then those candidates are re-evaluated by the relatively heavy RNN-LM.

\subsection{Korean Syllable and Tokenization}\label{ssec:korean}

We first introduce the basic structure of the Korean language.
Korean alphabet consists of 14 consonants and 10 vowels that form a Korean syllable.
A Korean syllable is divided into three parts: leading consonant (LC), vowel (V), and trailing consonant (TC) (see Figure~\ref{fig:korean}).
These three parts produce a unique sound, combined and pronounced together as a syllable.
Each part may use a single consonant/vowel or combination of consonants/vowels, resulting in 19 leading consonants, 21 vowels, and 27 trailing consonants.
In addition, silence can be used in place of LC and TC.
When the silence comes at LC position, a consonant `ㅇ' is explicitly written.
However, when the silence appears at TC, the position is left empty to indicate the silence.
Therefore, it is natural to choose LC, V, and TC as a minimum unit for Korean speech recognition.

Basically, four types of tokenization are commonly used for Korean \cite{korean_morpheme,korean_tokenization,korean_subword}: (1) word-based, (2) morpheme-based, (3) syllable-based and (4) consonant/vowel(phoneme)-based. 
For speech recognition, the former two tokenization methods may not be suitable; word-based tokenization suffers from the out-of-vocabulary problem and morpheme-based tokenization often does not match actual sound boundaries.
We also drop syllable-based tokenization because it requires a large token size; for example, the most common 2,306 syllables\footnote{The number of every possible Korean syllable combinations is\\ $19\times21\times(27+1)=11,172$.} are often considered~\cite{ksponspeech}, which may be too large for CTC-based speech recognition. 
For the consonant/vowel-based approach, we can either consider each LC, V, TC as a single token or further combine these tokens.
Based on the knowledge of native speakers, we decided to combine LC+V, because it can work as a single syllable independently.
LC+V or LC+V+TC can become a single syllable but LC or TC itself cannot behave as a syllable.
Therefore, we employ LC+V and TC separately as minimum units, for both ASR system and LM.

%% file: section/method.tex
\section{Proposed Korean Tokenization}\label{sec:method}


\begin{figure}[t]
    \centering
    \resizebox{1.0\linewidth}{!}{
    \includegraphics{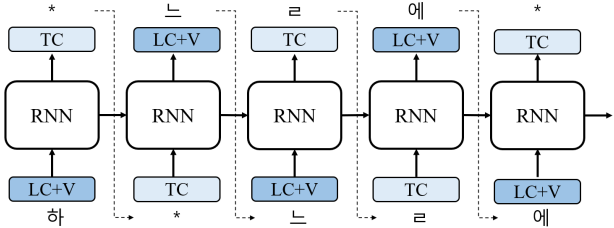}}
    \caption{Illustration of the input sequence for RNN-LM tokenized with the proposed SkipTC token (`` * ''). 
    The sequence becomes regularly patterned when SkipTC token is employed.}
    \label{fig:rnn}
    \vspace{-0.3cm}
\end{figure}

We propose SkipTC token for Korean tokenization to improve recognition performance.
Specifically, we tokenize each Korean syllable into two parts, (1) LC+V and (2) TC.
When TC is not included (skipped) in the syllable, we put an auxiliary token SkipTC (`` * '') in the place.
Table~\ref{tab:compare} compares various tokenization methods.
Our tokenization always provides a sequence of altering token types of LC+V and TC, including SkipTC token.
Although using SkipTC token increases the total sequence length, we claim that the difficulty of the probability density estimation can be decreased because of the regular input pattern (see Figure~\ref{fig:rnn}).
Intuitively, RNN-LM can be trained to predict the next token only among the correct tokens that are in turn.
We train RNN-LM using LSTM cells(thus, LSTM-LM) with the proposed tokenization, and then utilize the LM for the rescoring.
We measure the quality of LM in two ways; by the perplexity of LM itself and by the speech recognition error rate.

The proposed tokenization method has three advantages.
First, using SkipTC token helps RNN-LM to learn more regular patterns.
Second, the method only requires a vocabulary of 427 tokens: 399 LC+V tokens, 27 TC tokens, and 1 SkipTC token.
Third, our tokenization aligns well with the actual pronunciation that can be beneficial for both ASR and LM.

Algorithm~\ref{alg:scoring} shows the tokenization of a sentence and its LM likelihood score computation procedure.
We first insert SkipTC where the trailing consonant is missing, and then evaluate the extended sequence with LM trained with SkipTC token.
Note that $|Y|$ is not affected by inserting SkipTC, as we only count tokens that are included in ASR output vocabulary.
In the case of two-pass rescoring, the second term of Eq.~(\ref{eq:score}), $\log p_{\text{LM}}(Y)$, is replaced from the N-gram LM score to the RNN-LM score.

\begin{algorithm}[t]
\caption{LM Scoring with SkipTC tokenization}\label{alg:scoring}
\begin{algorithmic}[1]
\Procedure{LMScore}{$y_{1:i}$}
    \State $z = \{ y_{1} \}$
    \For{$j = \{2, ... i\}$}
        \If{($y_{j-1} \in$ LC+V) \textbf{and} ($y_{j} \in$ \{LC+V, space\})}
            \State $z \gets *$ \Comment{Append SkipTC token}
        \EndIf    
        \State $z \gets y_{j}$ \Comment{Append original token}
    \EndFor
    \State \textbf{return} $\log( p_{\text{LM}}( z ))$
\EndProcedure
\end{algorithmic}
\end{algorithm}
\vspace{0.2cm}




%% file: section/experiment.tex
\section{Experiment}\label{sec:experiment}

\subsection{Dataset}\label{ssec:dataset}

We evaluate the proposed method on the \textit{Korean Free Conversation} dataset~\cite{aihub}, a recently introduced large-scale Korean speech corpus.
Each utterance in the dataset is a recorded pronunciation of a colloquial sentence by native Korean speakers.
The dataset is composed of three categories (child, adult, senior) and two sub-categories (male, female) based on the age and gender of the speaker.
Table \ref{tab:data} shows the statistics of the dataset.
Because there is no specified test subset, we exploit the given valid dataset as a test dataset and split the training dataset (about 10\%) as a separate valid dataset.
For LM training, we employ the transcriptions from the same dataset as ASR; no external text source is used.

\begin{table}[h]
    \centering
    \vspace{0.1cm}
    \caption{Statistics of \textit{Korean Free Conversation} dataset.}
    \resizebox{0.88\linewidth}{!}{
    \begin{tabular}{c|rrr}
        \toprule
         & \#Utterances & \#Syllables & Hours(h) \\
        \midrule
        Train   & 4,000,000 & 74,797,987    & 6,089.2 \\  
        Valid   & 421,770   & 7,878,410     & 640.9 \\  
        Test    & 557,564   & 10,116,858    & 871.4 \\  
        \bottomrule
    \end{tabular}}
    \label{tab:data}
    \vspace{0.2cm}
\end{table}

We put a special effort into filtering and normalizing the data.
First, we remove the non-linguistic event symbols (i.e., noise, laughter) and punctuation marks (i.e., period, exclamation mark, question mark, quotation mark) so that the sentence only contains syllables.
Second, we filter out sentences consisting of less than 4 syllables.
Third, we remove audio files that are too short, corrupted, or not recorded at 16kHz.
After the cleaning, the average length and the number of syllables of each utterance are about 5.5 seconds and 18.6 syllables, respectively.

\subsection{Model and Training Setup}\label{ssec:setup}

We use Conformer-M~\cite{conformer} as the ASR encoder.
The input frames are 80-dim log-Mel filterbanks extracted from 25ms window and 15ms overlap.
We train the model with CTC loss using AdamW optimizer and peak learning rate of $1e^{-3}$.
During the training, we exploit SpecAugment~\cite{specaugment}, dropout of 0.1, and weight decay of $1e^{-5}$ for the regularization.
The learning rate linearly increases for 5K iterations and is fixed at the peak for 295K iterations.
Then, the learning rate decays following the inverse square root schedule~\cite{transformer} for 700K iterations.
The batch size is set to 16 for each GPU and 4x RTX Titan(24GB) GPUs are used.
Gradients are accumulated for 4 batches, expanding the effective batch size to 256.
We follow additional training details from previous work that employed the same encoder~\cite{shim}.

LSTM-LM and Transformer-LM are a stack of LSTM and Transformer-XL~\cite{TXL} layers, respectively.
We train these neural network-based LMs with and without SkipTC, sharing the same training configuration.
LMs consist of 4 layers where the hidden dimension of each layer is 512.
Vocabulary size is set to 431, including 399 LC+V tokens, 27 TC tokens and 5 special tokens: <pad>, <sos>, <eos>, <unk> and <space>.
When using SkipTC token, the <unk> token takes the role of SkipTC token.
The embedding weight and the final output weight are tied as the common practice~\cite{tying}. 
We train LSTM-LM using SGD with a momentum of 0.9, an initial learning rate of 0.1, weight decay of $1e^{-6}$, and exponential learning rate decay with a factor of 0.99.
For Transformer-LM, we use Adam optimizer with a learning rate of 0.001.
The training is conducted for 50 epochs with a batch size of 128.

\subsection{LM Performance}\label{ssec:lm}

\begin{table}[t]
    \centering
    \caption{LM performance comparison. Negative log-likelihood (\textit{nll}) values are reported.}
    \resizebox{0.96\linewidth}{!}{
    \begin{tabular}{cc|cc}
        \toprule
        LM type & SkipTC & Valid \textit{nll} $\downarrow$ & Test \textit{nll} $\downarrow$ \\
        \midrule
        \midrule
        4-gram LM & \xmark       & 1.865 & 1.956 \\
        6-gram LM & \xmark       & 1.189 & 1.461 \\
        \midrule
        LSTM-LM & \xmark     & 0.807 & 1.037 \\
        LSTM-LM & \cmark     & \textbf{0.598} & \textbf{0.775} \\
        \midrule
        Transformer-LM & \xmark     & 0.794 & 1.061 \\
        Transformer-LM & \cmark     & 0.604 & 0.803 \\
        \bottomrule
    \end{tabular}}
    \label{tab:nll}
\end{table}

To demonstrate the effectiveness of the proposed SkipTC, we train N-gram LMs, LSTM-LMs, and Transformer-LMs on the same training dataset.
Table~\ref{tab:nll} compares the LM performance between different LM types.
The result shows that utilizing SkipTC provides a considerable gain in performance for both LSTM-LM and Transformer-LM.
Specifically, using SkipTC reduces the negative log-likelihood value by $0.21$ and $0.26$ on the valid and test datasets, respectively.
N-gram models show the worst performance as expected.
We observe that the difference in language modeling performance is marginal for LSTM-LM and Transformer-LM; therefore, we decide to employ LSTM-LM for the rescoring.
The result empirically supports our speculation that patterning the input can be beneficial for neural network-based LMs in estimating the probability of a given sentence.

\subsection{ASR Performance}\label{ssec:asr}

Table \ref{tab:wer_ser} shows the ASR performance using different LMs.
We evaluate 4-gram, 6-gram LMs~\cite{kenlm}, and LSTM-LMs with and without SkipTC token.
LSTM-LMs are employed for two-pass rescoring after the first pass decoding exploiting the 6-gram LM.
We use a beam width of 128 for all cases.
To search the best hyper-parameter $\alpha$ and $\beta$, we first perform a grid search of $\alpha \in \{0.2, 0.4, 0.6, 0.8\}$ while keep $\beta=0$.
After $\alpha$ is selected, we fix $\alpha$ and search $\beta \in \{0.0, 1.0, 2.0, 4.0\}$.
We test hyper-parameter choices on the validation dataset and apply the best $\alpha$, $\beta$ for the test dataset.

Without the external LM decoding, the baseline ASR model achieves 9.27\% WER and 2.19\% SER on the test dataset.
After the rescoring, LSTM-LM with SkipTC token reduces the WER to 6.59\% and SER to 1.46\%.
We observe that utilizing SkipTC token shows better performance than the counterpart without SkipTC token; our method further lowers WER of validation and test dataset by 0.19\% and 0.14\%, respectively.


\begin{table}[t]
    \centering
    \caption{ASR performance comparison using different LMs for the beam search decoding.}
    \begin{subtable}[t]{0.98\linewidth}
        \centering
        \caption{Word error rate (WER)}
        \begin{tabular}{l|cc}
            \toprule
             & Valid (\%) & Test (\%) \\
            \midrule
            \midrule
            No LM decoding         & 6.72 & 9.27 \\    
            \midrule
            4-gram LM              & 5.84 & 8.57 \\    
            6-gram LM              & 4.39 & 7.87 \\    
            \midrule
            LSTM-LM (w/o SkipTC)   & 3.94    & 6.73 \\    
            LSTM-LM (w/  SkipTC)   & \textbf{3.75}    & \textbf{6.59} \\  
            \bottomrule
        \end{tabular}
        \label{tab:wer}
    \end{subtable}
    \begin{subtable}[t]{0.98\linewidth}
        \centering
        \vspace{0.3cm}
        \caption{Syllable error rate (SER)}
        \begin{tabular}{l|cc}
            \toprule
             & Valid (\%) & Test (\%) \\
            \midrule
            \midrule
            No LM decoding         & 1.65 & 2.19 \\    
            \midrule
            4-gram LM              & 1.34 & 1.96 \\    
            6-gram LM              & 0.96 & 1.80 \\    
            \midrule
            LSTM-LM (w/o SkipTC)   & 0.87    & 1.50 \\  
            LSTM-LM (w/  SkipTC)   & \textbf{0.80}    & \textbf{1.46} \\  
            \bottomrule
        \end{tabular}
        \label{tab:ser}
    \end{subtable}
    \label{tab:wer_ser}
\end{table}


\subsection{Discussion and Future Work}\label{ssec:discussion}

There is a relatively big performance gap for WER than SER because many errors appear from the incorrect spacing, which is one of the most common errors in Korean ASR~\cite{ksponspeech}.
In addition, we observe that a large amount of palatalization, nasalization, and lateralization errors are significantly disappeared when SkipTC is employed.  
Please note that these errors frequently appear for native Korean young children (not used to writing), because dictating the utterance as-is would cause such errors, particularly related to understanding when to omit TC or not.

The core idea of this paper is to propose SkipTC token for the improved LM rescoring, but only (LC+V, TC) tokenization is utilized as a baseline.
It would be an interesting topic to study if SkipTC token can also benefit other tokenization methods, such as (LC, V, TC).
Since a standard method for Korean ASR has not yet been established, more research on tokenization should be conducted not only for LM but also for the ASR system.

%% file: section/conclusion.tex
\section{Conclusion}\label{sec:conclusion}

In this work, we proposed a Korean tokenization method for the RNN-LM used for ASR rescoring.
We introduced a new auxiliary token, named SkipTC token, which is inserted when the trailing consonant is missing in a Korean syllable.
As a result, the input sequence for the RNN-LM becomes highly ordered and predictable, which eases the difficulty of the LM objective to learn the linguistic pattern.
We showed that using SkipTC token improves not only the LM performance but also the ASR performance.
We evaluated LM and ASR performance on the recently released large-scale Korean Free Conversation dataset for the first time.